\documentclass[runningheads]{llncs}

\usepackage{graphicx,subcaption} \graphicspath{ {Images/} {Images/sup/} }
\usepackage{amsmath,amssymb}
\usepackage[utf8]{inputenc}
\usepackage[ruled]{algorithm2e}
\usepackage{booktabs}
\usepackage{microtype}
\usepackage[hidelinks]{hyperref}

\DeclareMathOperator{\divergence}{div}

\begin{document}

\title{End-to-End Trainable Deep Active Contour Models for Automated Image Segmentation:\\ Delineating Buildings in Aerial Imagery}
\titlerunning{Trainable Deep Active Contours}

\author{Ali Hatamizadeh \and Debleena Sengupta \and Demetri Terzopoulos}
\authorrunning{A.~Hatamizadeh, D.~Sengupta, and D.~Terzopoulos}

\institute{Computer Science Department \\ University of California, Los Angeles, CA 90095, USA\\
\email{\{ahatamiz,debleenas,dt\}@cs.ucla.edu}\\}

\maketitle

\begin{abstract}
The automated segmentation of buildings in remote sensing imagery is a
challenging task that requires the accurate delineation of multiple
building instances over typically large image areas. Manual methods
are often laborious and current deep-learning-based approaches fail to
delineate all building instances and do so with adequate accuracy. As
a solution, we present Trainable Deep Active Contours (TDACs), an
automatic image segmentation framework that intimately
unites Convolutional Neural Networks (CNNs) and Active Contour Models
(ACMs). The Eulerian energy functional of the ACM component includes
per-pixel parameter maps that are predicted by the backbone CNN, which
also initializes the ACM. Importantly, both the ACM and CNN components
are fully implemented in TensorFlow and the entire TDAC architecture
is end-to-end automatically differentiable and backpropagation
trainable without user intervention. TDAC yields fast, accurate, and
fully automatic simultaneous delineation of arbitrarily many buildings
in the image. We validate the model on two publicly available aerial
image datasets for building segmentation, and our results demonstrate
that TDAC establishes a new state-of-the-art performance.

\end{abstract}

\section{Introduction}

The delineation of buildings in remote sensing imagery
\cite{lillesand2015remote} is a crucial step in applications such as
urban planning~\cite{shrivastava2017remote}, land cover analysis
\cite{zhang2018urban}, and disaster relief response
\cite{rudner2019multi3net}, among others. Manual or semi-automated
approaches can be very slow, laborious, and sometimes imprecise,
which can be detrimental to the prompt, accurate extraction of
situational information from high-resolution aerial and satellite 
images.

Convolutional Neural Networks (CNNs) and deep learning have been
broadly applied to various computer vision tasks, including semantic
and instance segmentation of natural images in general
\cite{chen2018deeplab,chen2019hybrid} and particularly to the
segmentation of buildings in remote sensing imagery
\cite{rudner2019multi3net,bischke2019multi}. However, building
segmentation challenges CNNs. First, since CNN architectures often
include millions of trainable parameters, successful training relies
on large, accurately-annotated datasets; but creating such datasets from
high-resolution imagery with possibly many building
instances is very tedious. Second, CNNs rely on a filter learning
approach in which edge and texture features are learned together,
which adversely impacts the ability to properly delineate buildings
and capture the details of their boundaries
\cite{geirhos2018,hatamizadeh2019endbound}.

One of the most influential computer vision techniques, the Active
Contour Model (ACM)~\cite{kass1988snakes}, has been successfully
employed in various image analysis tasks, including segmentation. In
most ACM variants, the deformable curves of interest dynamically
evolve according to an iterative procedure that minimizes a
corresponding energy functional. Since the ACM is a model-based
formulation founded on geometric and physical principles, the
segmentation process relies mainly on the content of the image itself,
not on learning from large annotated image datasets with hours or
days of training and extensive computational resources. However, the
classic ACM relies to some degree on user input to specify the initial
contour and tune the parameters of the energy functional, which
undermines its usefulness in tasks requiring the automatic
segmentation of numerous images.

\begin{figure}[t] \centering
\includegraphics[width=\linewidth]{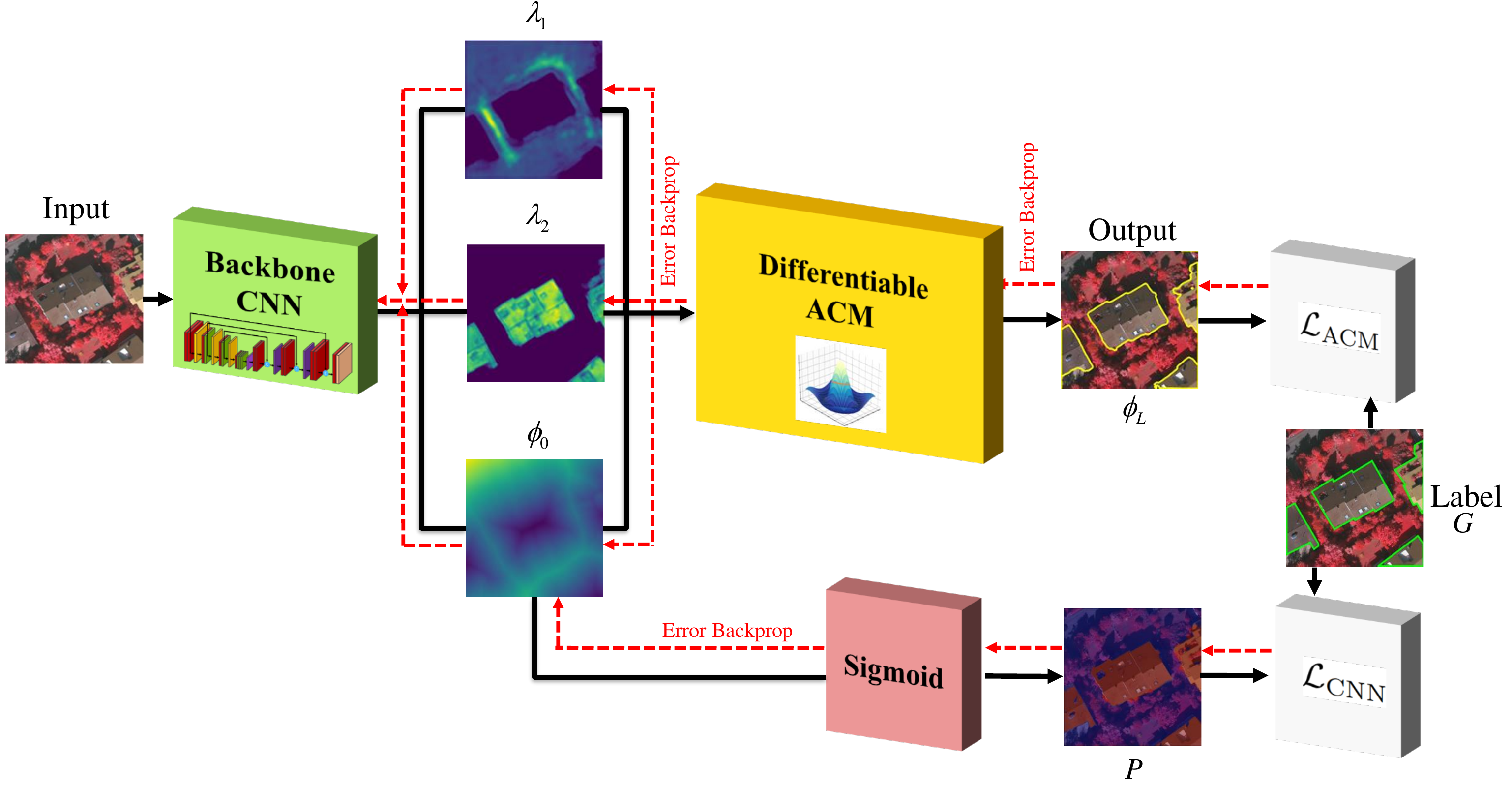}
\caption{TDAC is a fully-automated, end-to-end automatically differentiable and backpropagation trainable ACM and backbone CNN framework.}
\label{fig:framework}
\end{figure}

We address the aforementioned challenges by intimately uniting CNNs
and ACMs in an end-to-end trainable framework (originally proposed in
\cite{hatamizadeh2019endtoend}). Our framework (Fig.~\ref{fig:framework}) leverages a novel ACM
with trainable parameters that is automatically differentiable in a
TensorFlow implementation, thereby enabling the backpropagation of
gradients for stochastic optimization.
Consequently, the ACM and an untrained, as opposed to pre-trained,
backbone CNN can be trained together from scratch. Furthermore, our
ACM utilizes an Eulerian energy functional that affords local control
via 2D parameter maps that are directly predicted by the backbone CNN,
and it is also automatically initialized by the CNN. Thus, our
framework alleviates the biggest obstacle to exploiting the power of
ACMs in the context of CNNs and deep learning approaches to image
segmentation---eliminating the need for any form of user supervision
or intervention.

Our specific technical contributions in this paper are as follows:
\begin{itemize}
\item We propose an end-to-end trainable building segmentation
framework that establishes a tight merger between the ACM and any
backbone CNN in order to delineate buildings and accurately capture
the fine-grained details of their boundaries.

\item To this end, we devise an implicit ACM formulation with
pixel-wise parameter maps and differentiable contour propagation steps for
each term of the associated energy functional, thus making it amenable
to TensorFlow implementation.

\item We present new state-of-the-art benchmarks on two popular
publicly available building segmentation datasets, \textit{Vaihingen}
and \textit{Bing Huts}, with performance surpassing the best among
competing methods~\cite{marcos2018learning,cheng2019darnet}.
\end{itemize}

\section{Related Work}

\subsection{CNN-Based Building Segmentation Models}

Audebert \textit{et al.}~\cite{audebert2016semantic} leveraged
CNN-based models for building segmentation by applying
SegNet~\cite{badrinarayanan2017segnet} with multi-kernel convolutional
layers at three different resolutions. Subsequently, Wang \textit{et
al.}~\cite{wang2016torontocity} applied ResNet~\cite{he2016deep},
first to identify the instances, followed by an MRF to refine the
predicted masks. Some methods combine CNN-based models with classical
optimization methods. Costa \textit{et al.}~\cite{Costea_2017_ICCV}
proposed a two-stage model in which they detect roads and
intersections with a Dual-Hop Generative Adversarial Network (DH-GAN)
at the pixel level and then apply a smoothing-based graph optimization
to the pixel-wise segmentation to determine a best-covering road
graph. Wu \textit{et al.}~\cite{wu2018automatic} employed a U-Net
encoder-decoder architecture with loss layers at different scales to
progressively refine the segmentation masks. Xu \textit{et al.}
\cite{xu2018building} proposed a cascaded approach in which
pre-processed hand-crafted features are fed into a Residual U-Net to
extract building locations and a guided filter refines the results.

In an effort to address the problem of poor boundary predictions by
CNN models, Bischke \textit{et al.}~\cite{bischke2019multi} proposed a cascaded multi-task loss function to
simultaneously predict the semantic masks and distance classes.
Recently, Rudner \textit{et al.}~\cite{rudner2019multi3net} proposed a
method to segment flooded buildings using multiple streams of
encoder-decoder architectures that extract spatiotemporal information
from medium-resolution images and spatial information from
high-resolution images along with a context aggregation module to
effectively combine the learned feature map.

\subsection{CNN/ACM Hybrid Models}

Hu \textit{et al.}~\cite{hu2017deep} proposed a model in which the
network learns a level-set function for salient objects; however, the
authors predefined a fixed scalar weighting parameter $\lambda$, which
will not be optimal for all cases in the analyzed set of images.
Hatamizadeh \textit{et al.}~\cite{hatamizadeh2019deepactive} connected
the output of a CNN to an implicit ACM through spatially-varying functions
for the $\lambda$ parameters. Le \textit{et
al.}~\cite{le2018reformulating} proposed a framework for the task of
semantic segmentation of natural images in which level-set ACMs are
implemented as RNNs. There are three key differences between that effort
and our TDAC: (1) TDAC does not reformulate ACMs as RNNs, which makes
it more computationally efficient. (2) TDAC benefits from a novel,
locally-parameterized energy functional, as opposed to constant weighted
parameters (3) TDAC has an entirely different pipeline---we employ a
single CNN that is trained from scratch along with the ACM, as opposed
to requiring two pre-trained CNN backbones. The dependence
of~\cite{le2018reformulating} on pre-trained CNNs limits its
applicability.

Marcos \textit{et al.}~\cite{marcos2018learning} proposed Deep
Structured Active Contours (DSAC), an integration of ACMs with CNNs in
a structured prediction framework for building instance segmentation
in aerial images. There are three key differences between that work and
our TDAC: (1) TDAC is fully automated and runs without any external
supervision, as opposed to depending heavily on the manual
initialization of contours. (2) TDAC leverages the Eulerian ACM, which
naturally segments multiple building instances simultaneously, as
opposed to a Lagrangian formulation that can handle only a single
building at a time. (3) Our approach fully automates the direct
back-propagation of gradients through the entire TDAC framework due to
its automatically differentiable ACM implementation.

Cheng \textit{et al.}~\cite{cheng2019darnet} proposed the Deep Active
Ray Network (DarNet), which uses a polar coordinate ACM formulation to
prevent the problem of self-intersection and employs a computationally
expensive multiple initialization scheme to improve the performance of
the proposed model. Like DSAC, DarNet can handle only single instances
of buildings due to its explicit ACM formulation. Our approach is
fundamentally different from DarNet, as (1) it uses an implicit ACM
formulation that handles multiple building instances and (2) leverages
a CNN to automatically and precisely initialize the implicit ACM. 

Wang \textit{et al.}~\cite{wang2019object} proposed an interactive object
annotation framework for instance segmentation in which a backbone CNN
and user input guide the evolution of an implicit ACM. Recently, Gur
\textit{et al.}~\cite{gur2019end} introduced Active Contours via
Differentiable Rendering Network (ACDRNet) in which an explicit ACM is
represented by a ``neural renderer'' and a backbone encoder-decoder
U-Net predicts a shift map to evolve the contour via edge
displacement.

Some efforts have also focused on deriving new loss functions that are
inspired by ACM principles. Inspired by the global energy formulation
of~\cite{chan2001active}, Chen \textit{et al.}~\cite{chen2019learning}
proposed a supervised loss layer that incorporated area and size
information of the predicted masks during training of a CNN and
tackled a medical image segmentation task. Similarly, Gur \textit{et
al.}~\cite{gur2019unsupervised} presented an unsupervised loss
function based on morphological active contours without
edges~\cite{marquez2013morphological}.

\begin{figure}[t] \centering
\includegraphics[scale=0.33]{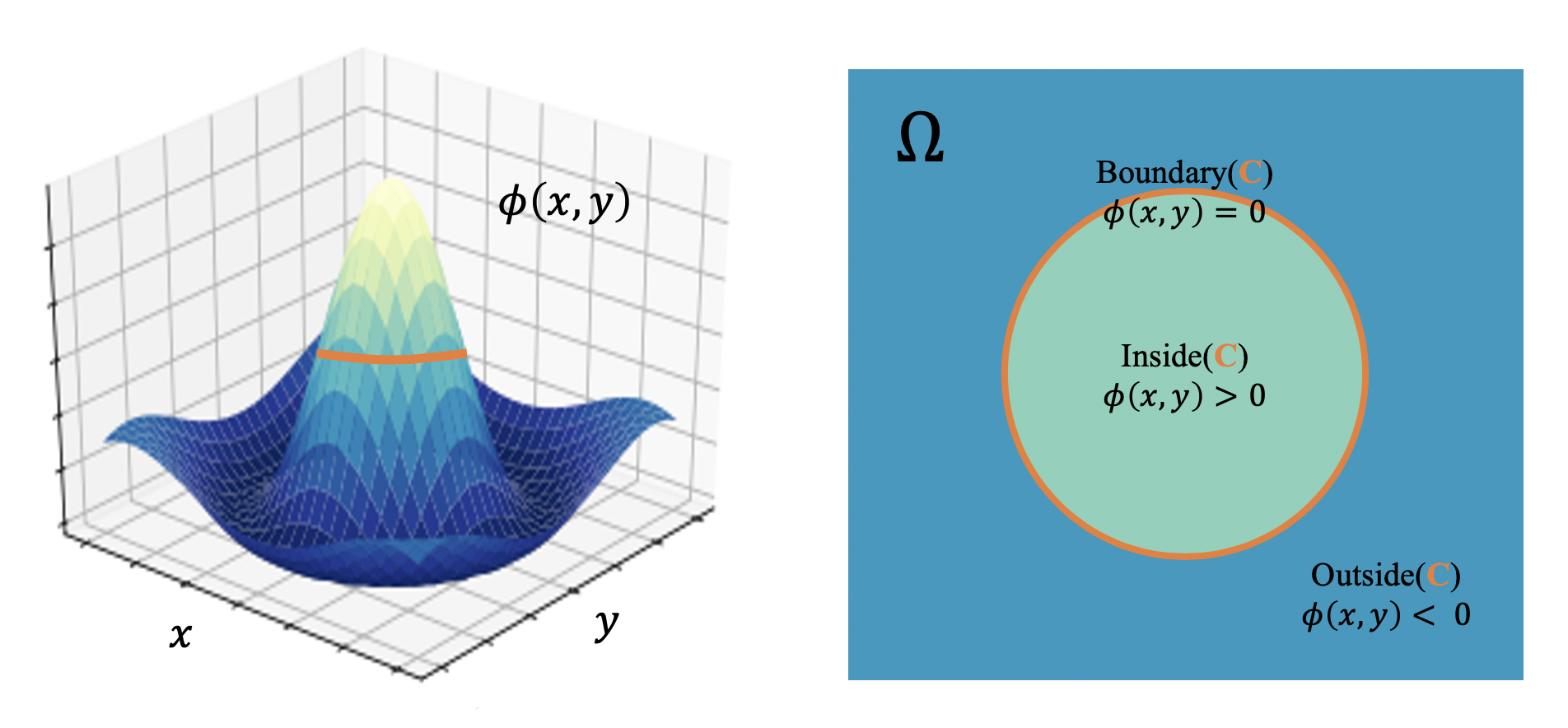} 
\caption{Boundary $C$ represented as the zero level-set of implicit function $\phi(x,y)$.}
\label{fig:level-set}
\end{figure}

\section{The TDAC Model}

\subsection{Localized Level-Set ACM with Trainable Parameters}

Our ACM formulation allows us to create a differentiable and trainable
active contour model. Instead of working with a parametric contour
that encloses the desired area to be segmented~\cite{kass1988snakes},
we represent the contour(s) as the zero level-set of an implicit
function. Such so-called ``level-set active contours'' evolve the
segmentation boundary by evolving the implicit function so as to
minimize an associated Eulerian energy functional.

The most well-known approaches that utilize this implicit formulation are
geodesic active contours \cite{caselles1997geodesic} and active
contours without edges \cite{chan2001active}. The latter, also known as the Chan-Vese
model, relies on image intensity differences between the interior and
exterior regions of the level set. Lankton and
Tannenbaum~\cite{lankton2008localizing} proposed a reformulation in
which the energy functional incorporates image properties in the local
region near the level set, which more accurately segments objects with
heterogeneous features.\footnote{These approaches numerically solve
the PDE that governs the evolution of the implicit function.
Interestingly, Márquez-Neila \textit{et
al.}~\cite{marquez2013morphological} proposed a morphological approach
that approximates the numerical solution of the PDE by successive
application of morphological operators defined on the equivalent
binary level set.}

\begin{figure}[t] \centering
\def\x{0.48}
\subcaptionbox{}{\includegraphics[width=\x\linewidth]{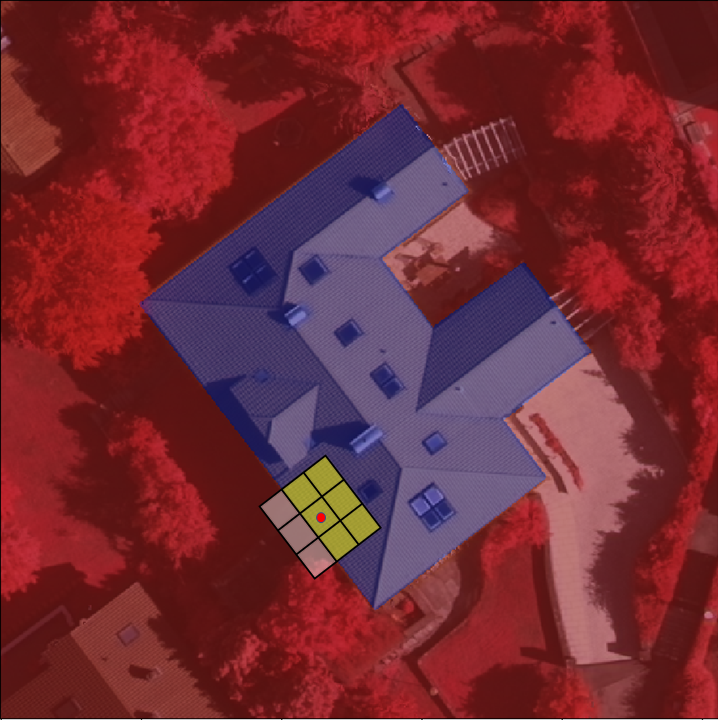}}
\hfill
\subcaptionbox{}{\includegraphics[width=\x\linewidth]{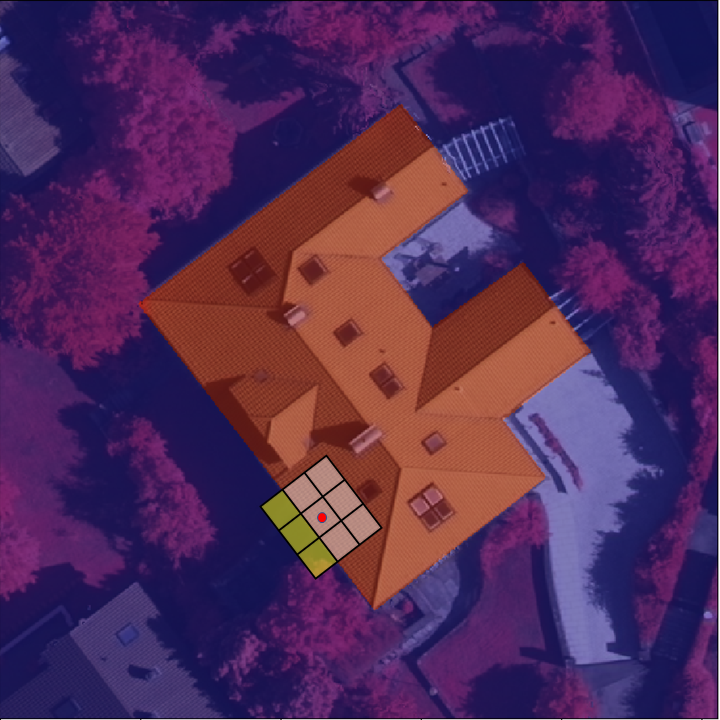}}
\caption{The filter is divided by the contour into interior and
exterior regions. The point $x$ is represented by the red dot and the
interior (a) and exterior (b) regions are shaded yellow.}
\label{fig:parameter_maps}
\end{figure}

Let $I$ represent an input image and $C= \big\{(x, y) | \phi(x, y) = 0
\big\}$ be a closed contour in $\Omega\in R^2$ represented by the zero
level set of the signed distance map $\phi(x,y)$
(Fig.~\ref{fig:level-set}). The interior and exterior of $C$ are
represented by $\phi(x,y)>0$ and $\phi(x,y)<0$, respectively.
Following~\cite{chan2001active}, we use a smoothed Heaviside function
\begin{equation}
H(\phi(x, y))=\frac{1}{2}+\frac{1}{\pi}\arctan\Big(
\frac{{\phi(x,y)}}{\epsilon}\Big)
\label{eq:heavi}
\end{equation}
to represent the interior as $H(\phi)$ and exterior as $(1-H(\phi))$.
The derivative of $H(\phi(x, y))$ is
\begin{equation}
\frac{\partial H(\phi(x, y))}{\partial
\phi(x,y)}=\frac{1}{\pi}\frac{\epsilon}{\epsilon^2+\phi(x,y)^2}
= \delta(\phi(x, y)).
\end{equation}
In TDAC, we evolve $C$ to minimize an energy function according to
\begin{equation}
E({\phi})= E_\textrm{length}({\phi})+E_\textrm{image}({\phi}),
\label{eq:contourloss1}
\end{equation}
where
\begin{equation}
E_\textrm{length}({\phi})= \int_\Omega
\mu\delta(\phi(x,y))|\nabla\phi(x,y)|\,dx\,dy
\label{eq:contourloss2}
\end{equation}
penalizes the length of $C$ whereas
\begin{equation}
\label{eq:uniform_density}
\begin{split}
E_\textrm{image}(\phi) &= \int_\Omega\delta(\phi(x,y))
\biggl[H(\phi(x,y))(I(x,y)-m_1)^2+\\
&\qquad\qquad(1-H(\phi(x,y)))(I(x,y)-m_2)^2\biggr]\,dx\,dy
\end{split}
\end{equation}
takes into account the mean image intensities $m_1$ and $m_2$ of the
regions interior and exterior to $C$~\cite{chan2001active}.
We compute these local statistics using a characteristic function
$W_{s}$ with local window of size
$f_{s}$ (Fig.~\ref{fig:parameter_maps}), as follows:
\begin{equation}
\begin{split}
\label{eq:charf}
& W_{s} =
  \begin{cases}
    1      & \quad \text{if } x-f_{s}\leq u \leq x+f_{s}, \quad y-f_{s}\leq v \leq y+f_{s};\\
    0      & \quad \text{otherwise},
  \end{cases}
\end{split}
\end{equation} 
where $x,y$ and $u,v$ are the coordinates of two independent points.

To make our level-set ACM trainable, we associate parameter maps with the foreground and
background energies. These maps, $\lambda_1(x,y)$ and $\lambda_2(x,y)$, are functions over the image
domain $\Omega$. Therefore, our energy function may be written as
\begin{equation}
\label{eq:f_pc1}
\begin{split}
E(\phi)= & \int_\Omega \delta(\phi(x,y)) \biggl[\mu|\nabla\phi(x,y)| +
\int_\Omega W_s F(\phi(u,v)) \,du\,dv\biggr]\,dx\,dy,
\end{split}
\end{equation}
where
\begin{equation}
\label{eq:density}
\begin{split}
F(\phi) &= \lambda_1(x,y) (I(u,v)-m_1(x,y))^2 (H(\phi(x,y))\\ &+
\lambda_2(x,y) (I(u,v)-m_2(x,y))^2 (1-H(\phi(x,y)).
\end{split}
\end{equation}
According to the derivation in the appendix, the variational
derivative of $E$ with respect to $\phi$ yields the Euler-Lagrange PDE
\begin{equation}
\frac{\partial\phi}{\partial t} = \delta(\phi) \biggl[\mu \divergence
\left(\frac{\nabla\phi}{|\nabla\phi|}\right) + \int_\Omega W_s
\nabla_\phi F(\phi)\,dx\,dy \biggr]
\label{eq:ACWEalgorithm}
\end{equation}
with
\begin{equation}
\label{eq:final_evol}
\nabla_\phi F = \delta(\phi) \bigl(\lambda_1(x,y)(I(u,v)-m_1(x,y))^2-
\lambda_2(x,y)(I(u,v)-m_2(x,y))^2\bigr).
\end{equation}
To avoid numerical
instabilities during the evolution and maintain a well-behaved
$\phi(x,y)$, a distance regularization term~\cite{li2010distance} can
be added to (\ref{eq:ACWEalgorithm}).

It is important to note that our formulation enables us to capture the
fine-grained details of boundaries, and our use of \emph{pixel-wise} parameter maps $\lambda_1(x,y)$ and $\lambda_2(x,y)$ allows them to be
directly predicted by the backbone CNN along with an initialization
map $\phi_0(x,y)$. Thus, not only does the implicit ACM propagation
now become fully automated, but it can also be directly controlled by
a CNN through these learnable parameter maps.

\begin{figure}[t] \centering
\includegraphics[width=\linewidth]{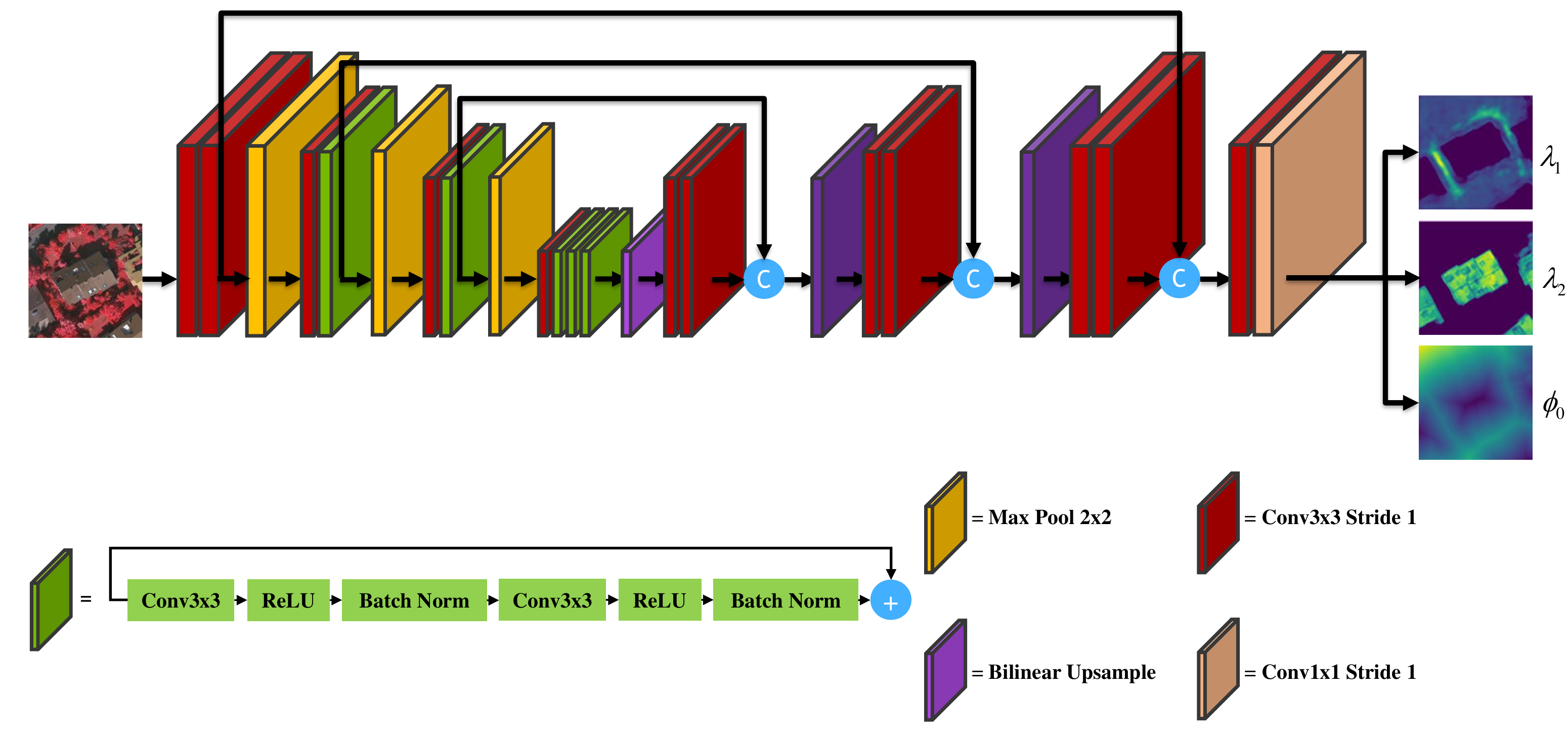}
\caption{TDAC's CNN backbone employs a standard encoder-decoder architecture.}
\label{fig:cnn}
\end{figure}

\subsection{CNN Backbone}

For the backbone CNN, we use a standard encoder-decoder with convolutional layers, residual blocks, and skip connections between the encoder and decoder. Each
$3\times3$ convolutional layer is followed by ReLU activation and
batch normalization. Each residual block consists of two $3\times3$
convolutional layers and an additive identity skip connection.
As illustrated in Fig.~\ref{fig:cnn}, the first stage of the encoder
comprises two $3\times3$ convolutional layers and a max pooling
operation. Its second and third stages are comprised of
a residual block followed by a max pooling operation. Each stage of
the decoder performs a bilinear upsampling followed by two
convolutional layers. The encoder is connected to the decoder via
three residual blocks as well as skip connections at every stage. The
output of the decoder is connected to a $1\times1$ convolution with
three output channels for predicting the $\lambda_{1}(x,y)$ and
$\lambda_{2}(x,y)$ parameter maps as well as the initialization map
$\phi_0(x,y)$.

\begin{table}[t] \centering
\setlength{\tabcolsep}{4pt}
\caption{Detailed information about the TDAC encoder.}
\label{tab:encoder}
\begin{tabular}{ll}
\toprule
Operations & Output size    \\ 
\midrule
Input & $512\times512\times3$   \\
Conv, ReLU, BN, Conv, ReLU, BN, Pool & $256\times256\times16$    \\
Conv, ReLU, BN & $256\times256\times32$    \\
Conv, ReLU, BN, Conv, ReLU, BN, Add, Pool & $128\times128\times32$    \\
Conv, ReLU, BN & $128\times128\times64$    \\
Conv, ReLU, BN, Conv, ReLU, BN, Add, Pool & $64\times64\times64$    \\
Conv, ReLU, BN & $64\times64\times128$    \\
Conv, ReLU, BN, Conv, ReLU, BN, Add & $64\times64\times128$    \\
Conv, ReLU, BN, Conv, ReLU, BN, Add & $64\times64\times128$    \\
Conv, ReLU, BN, Conv, ReLU, BN, Add & $64\times64\times128$    \\
\bottomrule
\end{tabular}
\end{table}

\begin{table}[t] \centering
\setlength{\tabcolsep}{4pt}
\caption{Detailed information about the TDAC decoder.}
\label{tab:decoder}
\begin{tabular}{ll}
\toprule
Operations & Output size    \\ 
\midrule
Input & $64\times64\times128$   \\
Upsample, Conv, ReLU, BN, Conv, ReLU, BN & $128\times128\times64$    \\
Upsample, Conv, ReLU, BN, Conv, ReLU, BN & $256\times256\times32$    \\
Upsample, Conv, ReLU, BN, Conv, ReLU, BN & $512\times512\times16$    \\
Conv, ReLu, BN & $512\times512\times16$ \\
Conv1 & $512\times512\times3$    \\
\bottomrule
\end{tabular}
\end{table}

\clearpage

Tables~\ref{tab:encoder} and \ref{tab:decoder} present the details of
the encoder and decoder in the TDAC backbone CNN architecture. BN,
Add, Pool, Upsample, Conv, and Conv1 denote batch normalization,
addition, $2\times2$ max pooling, bilinear upsampling, $3\times3$
convolutional, and $1\times1$ convolutional layers, respectively.

\subsection{Differentiable ACM}
\label{sec:diff_acm}

The ACM is evolved according to (\ref{eq:ACWEalgorithm}) in a
differentiable manner in TensorFlow. The first term is computed
according to the surface curvature expression:
\begin{equation}
\label{eq:f_pc}
\divergence \left(\frac{\nabla\phi}{|\nabla\phi|}\right) = \dfrac{\phi_{xx}\phi_{y}^{2}-2\phi_{xy}\phi_{x}\phi_{y}+\phi_{yy}\phi_{x}^{2}}{(\phi_{x}^2+\phi_{y}^{2})^{3/2}},
\end{equation}
where the subscripts denote the spatial partial derivatives of $\phi$,
which are approximated using central finite differences. For the second
term, convolutional operations are leveraged to
efficiently compute $m_{1}(x,y)$ and $m_{2}(x,y)$ in (\ref{eq:density}) within image regions interior and exterior to $C$. Finally,
${\partial\phi}/{\partial t}$ in (\ref{eq:ACWEalgorithm}) is evaluated and
$\phi(x,y)$ updated according to
\begin{equation}
\phi^{t}= \phi^{t-1} +\Delta t \frac{\partial\phi^{t-1}}{\partial t},
\label{eq:attloss1}
\end{equation}
where $\Delta t$ is the size of the time step.

\subsection{TDAC Training}
\label{sec:loss}

Referring to Fig.~\ref{fig:framework}, we simultaneously train the CNN and level-set components of TDAC in an
end-to-end manner with no human intervention. The CNN guides the ACM
by predicting the $\lambda_{1}(x,y)$ and $\lambda_{2}(x,y)$ parameter
maps, as well as an initialization map $\phi_0(x,y)$ from which $\phi(x,y)$
evolves through the $L$ layers of the ACM in a differentiable manner, thus enabling training error
backpropagation. The $\phi_0(x,y)$ output of the CNN is also passed
into a Sigmoid activation function to produce the prediction $P$.
Training optimizes a loss function that combines binary cross entropy and Dice losses:
\begin{equation}
\hat{\mathcal{L}}(X)= -\frac{1}{N}\sum_{j=1}^{N}\left[X_{j}\log G_{j}+(1-X_{j})\log(1-G_{j}) \right]+1-\frac{\sum_{j=1}^{N} 2
X_{j}G_{j}}{\sum_{j=1}^{N} X_{j} + \sum_{j=1}^{N}G_{j}},
\label{eq:ce}
\end{equation}
where $X_{j}$ denotes the output prediction and $G_{j}$ the
corresponding ground truth at pixel $j$, and $N$ is the total number
of pixels in the image. The total loss of the TDAC model is
\begin{equation}
\mathcal{L} = \mathcal{L}_\text{ACM} + \mathcal{L}_\text{CNN},
\label{eq:attloss}
\end{equation}
where $\mathcal{L}_\textrm{ACM}= \hat{\mathcal{L}}(\phi_L)$ is the
loss computed for the output $\phi_L$ from the final ACM layer and
$\mathcal{L}_\text{CNN}= \hat{\mathcal{L}}(P)$ is the loss computed
over the prediction $P$ of the backbone CNN. Algorithm~\ref{algorithm}
presents the details of the TDAC training procedure.

\begin{algorithm}[t]
 \caption{TDAC Training Algorithm}
 \KwData{$I$: Image; $G$: Corresponding ground truth label;
 $g$: ACM energy function with parameter maps $\lambda_1,\lambda_2$;
 $\phi$: ACM implicit function;
 $L$: Number of ACM iterations;
 $W$: CNN with weights $w$;
 $P$: CNN prediction;
 $\mathcal{L}$: Total loss function;
 $\eta$: Learning rate}
 \KwResult{Trained TDAC model}
 \While{not converged}{
  $\lambda_1,\lambda_2,\phi_{0}=W(I)$\\
  $P=\hbox{Sigmoid}(\phi_{0})$\\
 \For{$t=1$ \textbf{to} $L$}{
    $ \frac{\partial\phi_{t-1}}{\partial t}=g(\phi_{t-1};\lambda_1,\lambda_2, I)$\\
    $\phi^{t}= \phi^{t-1} +\Delta t \frac{\partial\phi^{t-1}}{\partial t}$
    }
    $\mathcal{L}=\mathcal{L}_\text{ACM}(\phi_{L})+\mathcal{L}_\text{CNN}(P)$\\
  Compute $\frac{\partial \mathcal{L}}{\partial w}$ and backpropagate the error\\
  Update the weights of $W$: $w \leftarrow w-\eta  \frac{\partial \mathcal{L}}{\partial w}$
}
\label{algorithm}
\end{algorithm}

\subsection{Implementation Details}

We have implemented the TDAC architecture and training algorithm
entirely in TensorFlow. Our ACM implementation benefits from the
automatic differentiation utility of TensorFlow and has been designed
to enable the backpropagation of the error gradient through the $L$ layers
of the ACM. We set $L=60$ iterations in the ACM
component of TDAC since, as will be discussed in
Section~\ref{sec:ablation}, the performance does not seem to improve
significantly with additional iterations. We set a filter size of
$f=5$, as discussed in Section~\ref{sec:ablation}. The training was
performed on an Nvidia Titan RTX GPU, and an Intel® Core™ i7-7700K CPU
@ 4.20GHz. The size of the training minibatches for both datasets is
2. All the training sessions employ the Adam optimization
algorithm~\cite{kingma2014adam} with an initial learning rate of
$\alpha_{0}=0.001$ decreasing according to~\cite{myronenko2019robust}
\begin{equation}
\label{eq:learningrate}
\alpha = \alpha_{0}\left(1-e/N_{e}\right)^{0.9} 
\end{equation}
with epoch counter $e$ and total number of epochs $N_{e}$.

\section{Empirical Study}
\label{sec:exp}

\subsection{Datasets}

\paragraph{Vaihingen:}
The Vaihingen buildings dataset consists of 168 aerial images of
size $512\times512$ pixels. Labels for each image were generated by
using a semi-automated approach. We used 100 images for training and
68 for testing, following the same data partition as
in~\cite{marcos2018learning}. In this dataset, almost all the images
include multiple instances of buildings, some of which are located at
image borders.

\paragraph{Bing Huts:}
The Bing Huts dataset consists of 605 aerial images of size $64\times64$
pixels. We followed the same data partition used
in~\cite{marcos2018learning}, employing 335 images for training and
270 images for testing. This dataset is especially challenging due the
low spatial resolution and contrast of the images.

\subsection{Evaluation Metrics}

To evaluate TDAC's performance, we utilized four different
metrics---Dice, mean Intersection over Union (mIoU), Boundary F
(BoundF)~\cite{cheng2019darnet}, and Weighted Coverage
(WCov)~\cite{w_cov}.

Given the prediction $X$ and ground truth mask $G$, the Dice (F1) score is
\begin{equation}
\textrm{Dice}(X,G) = \frac{2 \sum_{i=1}^{N}X_{i}G_{i} }{\sum_{i=1}^{N}X_{i} + \sum_{i=1}^{N} G_{i}},
\label{eq:dice}
\end{equation}
where $N$ is the number of image pixels and $G_{i}$ and $X_{i}$ denote
pixels in $G$ and $X$.

Similarly, the IoU score measures the overlap of two objects by
calculating the ratio of intersection over union, according to
\begin{equation}
\textrm{IoU}(X,G) = \frac{|X \cap G|}{|X \cup G|}.
\label{eq:iou}
\end{equation} 

BoundF computes the average of Dice scores over 1 to 5 pixels around
the boundaries of the ground truth segmentation.

In WConv, the maximum overlap output is selected and the IoU between
the ground truth segmentation and best output is calculated. IoUs for
all instances are summed up and weighted by the area of the ground
truth instance. Assuming that $S_{G} =
\{r_{1}^{S_{G}},\dots,r_{|S_{G}|}^{S_{G}}\}$ is a set of ground truth
regions and $S_{X} = \{r_{1}^{S_{X}},\dots,r_{|S_{X}|}^{S_{X}}\}$ is a
set of prediction regions for single image, and $|r_{j}^{S_{G}}|$ is
the number of pixels in $r_{j}^{S_{G}}$, the weighted coverage can be
expressed as
\begin{equation}
\textrm{WCov}(S_{X},S_{G}) = \frac{1}{N}\sum_{j=1}^{|S_{G}|}{|r_{j}^{S_{G}}|} \max\limits_{k=1...|S_{X}|} \textrm{IoU}(r_{k}^{S_{X}}, r_{j}^{S_{G}}).
\label{eq:wcov}
\end{equation} 

\subsection{Experiments and Ablation Studies}
\label{sec:ablation}

\begin{figure} \centering
\def\x{0.135}
\includegraphics[width=\x\linewidth,height=\x\linewidth]{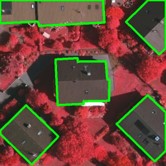}
\hfill
\includegraphics[width=\x\linewidth,height=\x\linewidth]{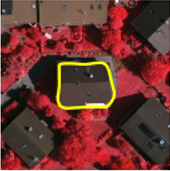}
\hfill
\includegraphics[width=\x\linewidth,height=\x\linewidth]{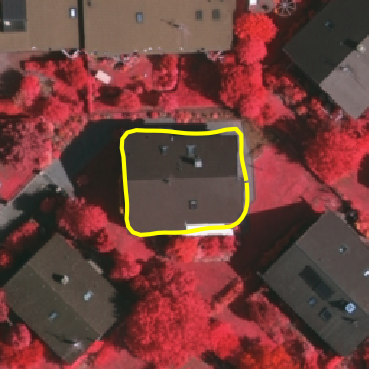}
\hfill
\includegraphics[width=\x\linewidth,height=\x\linewidth]{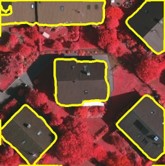}
\hfill
\includegraphics[width=\x\linewidth,height=\x\linewidth,trim={80
35 20 20},clip]{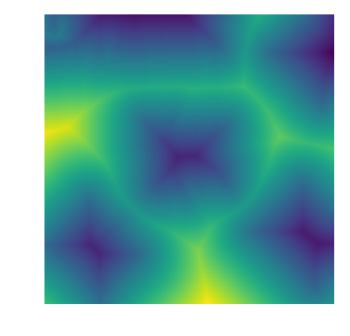} \hfill
\includegraphics[width=\x\linewidth,height=\x\linewidth,trim={80
35 20 20},clip]{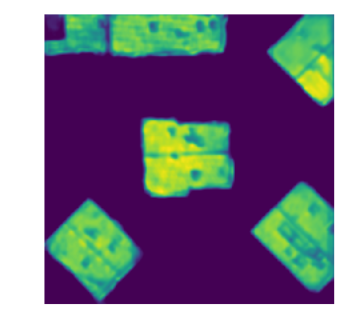} \hfill
\includegraphics[width=\x\linewidth,height=\x\linewidth,trim={80
35 20 20},clip]{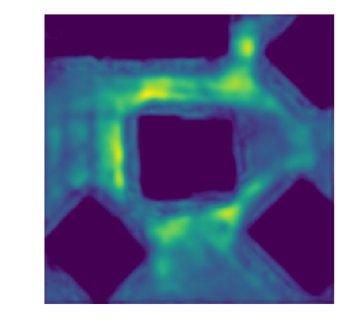}\\[4pt]

\includegraphics[width=\x\linewidth,height=\x\linewidth]{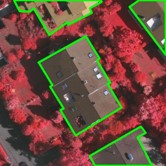}
\hfill
\includegraphics[width=\x\linewidth,height=\x\linewidth]{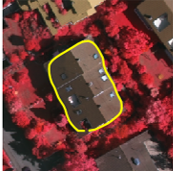}
\hfill
\includegraphics[width=\x\linewidth,height=\x\linewidth]{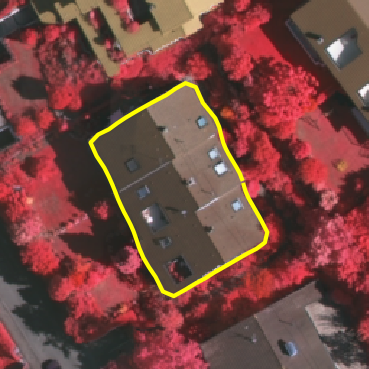}
\hfill
\includegraphics[width=\x\linewidth,height=\x\linewidth]{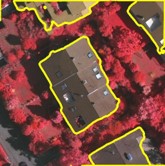}
\hfill
\includegraphics[width=\x\linewidth,height=\x\linewidth,trim={80
35 20 20},clip]{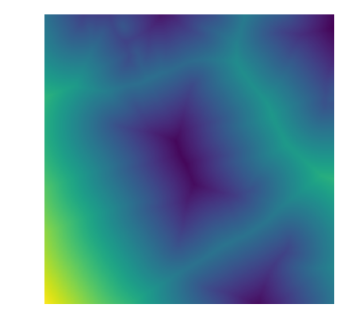} \hfill
\includegraphics[width=\x\linewidth,height=\x\linewidth,trim={80
35 20 20},clip]{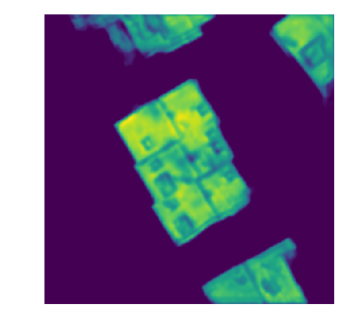} \hfill
\includegraphics[width=\x\linewidth,height=\x\linewidth,trim={80
35 20 20},clip]{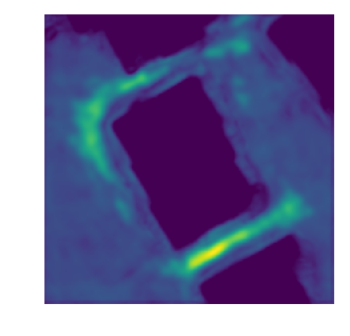}\\[4pt]

\includegraphics[width=\x\linewidth,height=\x\linewidth]{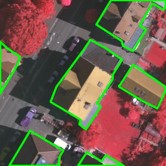}
\hfill
\includegraphics[width=\x\linewidth,height=\x\linewidth]{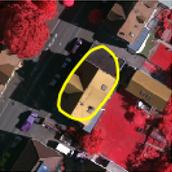}
\hfill
\includegraphics[width=\x\linewidth,height=\x\linewidth]{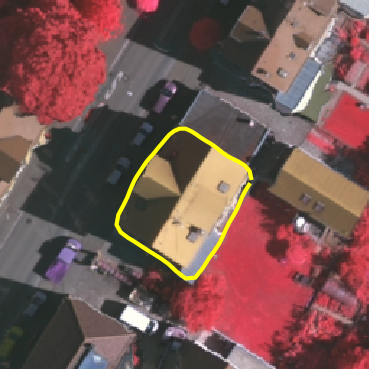}
\hfill
\includegraphics[width=\x\linewidth,height=\x\linewidth]{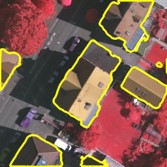}
\hfill
\includegraphics[width=\x\linewidth,height=\x\linewidth,trim={80
35 20 20},clip]{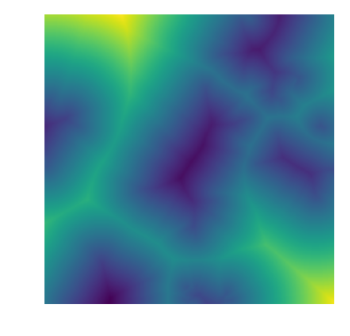} \hfill
\includegraphics[width=\x\linewidth,height=\x\linewidth,trim={80
35 20 20},clip]{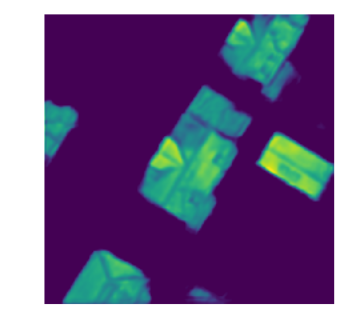} \hfill
\includegraphics[width=\x\linewidth,height=\x\linewidth,trim={80
35 20 20},clip]{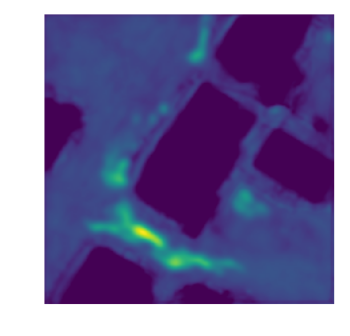}\\[4pt]

\includegraphics[width=\x\linewidth,height=\x\linewidth]{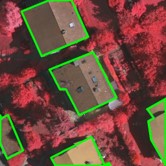}
\hfill
\includegraphics[width=\x\linewidth,height=\x\linewidth]{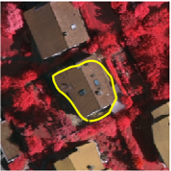}
\hfill
\includegraphics[width=\x\linewidth,height=\x\linewidth]{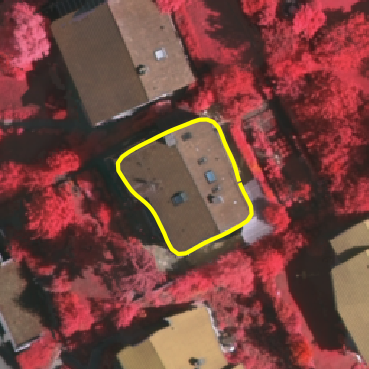}
\hfill
\includegraphics[width=\x\linewidth,height=\x\linewidth]{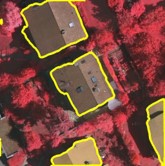}
\hfill
\includegraphics[width=\x\linewidth,height=\x\linewidth,trim={80
35 20 20},clip]{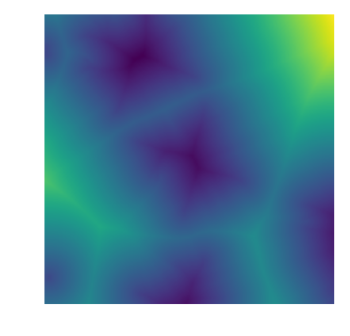} \hfill
\includegraphics[width=\x\linewidth,height=\x\linewidth,trim={80
35 20 20},clip]{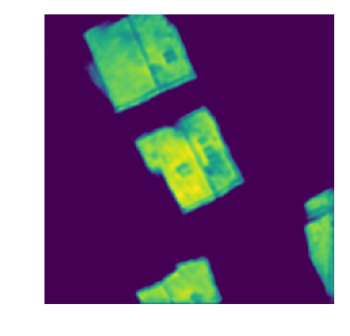} \hfill
\includegraphics[width=\x\linewidth,height=\x\linewidth,trim={80
35 20 20},clip]{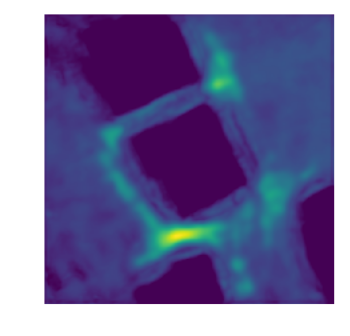}\\[4pt]
\hfill
\includegraphics[width=\x\linewidth,height=\x\linewidth]{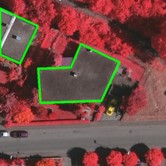}
\hfill
\includegraphics[width=\x\linewidth,height=\x\linewidth]{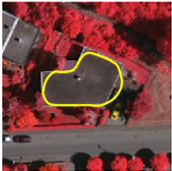}
\hfill
\includegraphics[width=\x\linewidth,height=\x\linewidth]{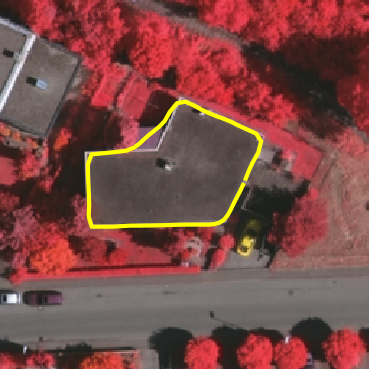}
\hfill
\includegraphics[width=\x\linewidth,height=\x\linewidth]{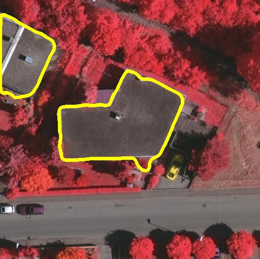}
\hfill
\includegraphics[width=\x\linewidth,height=\x\linewidth]{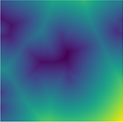} \hfill
\includegraphics[width=\x\linewidth,height=\x\linewidth]{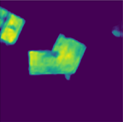} \hfill
\includegraphics[width=\x\linewidth,height=\x\linewidth]{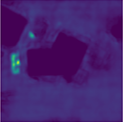}\\[8pt]

\includegraphics[width=\x\linewidth,height=\x\linewidth]{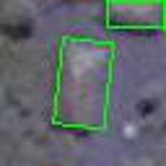}
\hfill
\includegraphics[width=\x\linewidth,height=\x\linewidth]{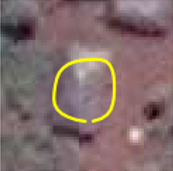}
\hfill
\includegraphics[width=\x\linewidth,height=\x\linewidth]{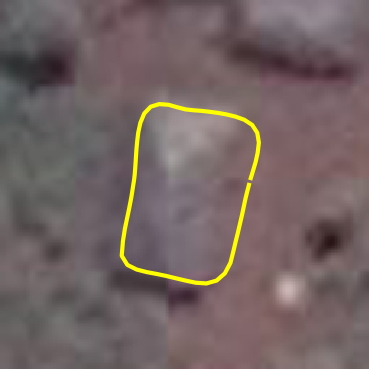}
\hfill
\includegraphics[width=\x\linewidth,height=\x\linewidth]{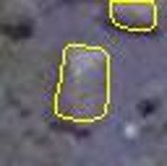}
\hfill
\includegraphics[width=\x\linewidth,height=\x\linewidth]{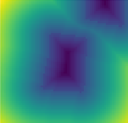}
\hfill
\includegraphics[width=\x\linewidth,height=\x\linewidth]{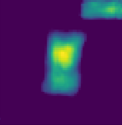}
\hfill
\includegraphics[width=\x\linewidth,height=\x\linewidth]{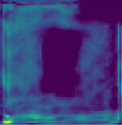}\\[4pt]

\includegraphics[width=\x\linewidth,height=\x\linewidth]{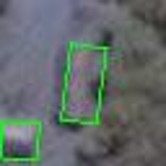}
\hfill
\includegraphics[width=\x\linewidth,height=\x\linewidth]{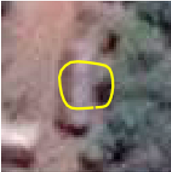}
\hfill
\includegraphics[width=\x\linewidth,height=\x\linewidth]{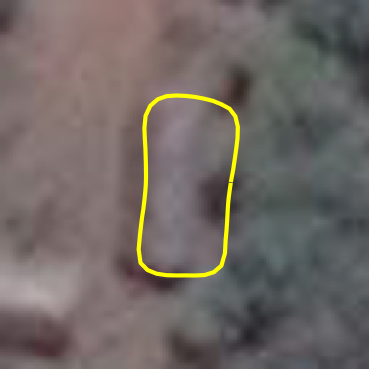}
\hfill
\includegraphics[width=\x\linewidth,height=\x\linewidth]{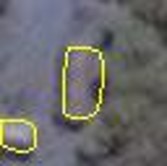}
\hfill
\includegraphics[width=\x\linewidth,height=\x\linewidth]{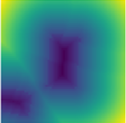}
\hfill
\includegraphics[width=\x\linewidth,height=\x\linewidth]{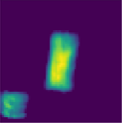}
\hfill
\includegraphics[width=\x\linewidth,height=\x\linewidth]{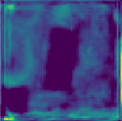}\\[4pt]

\begin{subfigure}{\textwidth}
  \centering
  \subcaptionbox{Image}{\includegraphics[width=\x\linewidth,height=\x\linewidth]{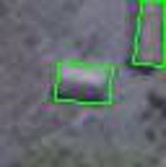}}
\hfill
\subcaptionbox{DSAC}{\includegraphics[width=\x\linewidth,height=\x\linewidth]{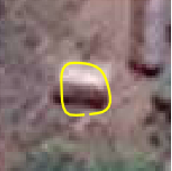}}
\hfill
\subcaptionbox{DarNet}{\includegraphics[width=\x\linewidth,height=\x\linewidth]{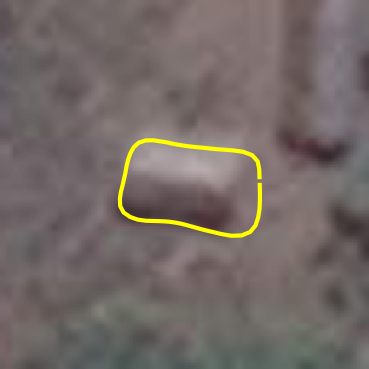}}
\hfill
\subcaptionbox{\bf TDAC}{\includegraphics[width=\x\linewidth,height=\x\linewidth]{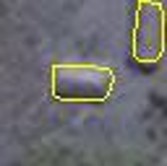}}
\hfill
\subcaptionbox{$\phi_0$}{\includegraphics[width=\x\linewidth,height=\x\linewidth]{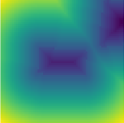}}
\hfill
\subcaptionbox{$\lambda_1$}{\includegraphics[width=\x\linewidth,height=\x\linewidth]{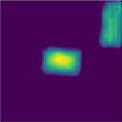}}
\hfill
\subcaptionbox{$\lambda_2$}{\includegraphics[width=\x\linewidth,height=\x\linewidth]{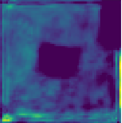}}\\[4pt]
\end{subfigure}
\caption{Comparative visualization of the labeled image and the outputs of
DSAC, DarNet, and our TDAC for the
Vaihingen (top) and Bing Huts (bottom) datasets. (a) Image labeled
with (green) ground truth segmentation. (b) DSAC output. (c) DarNet
output. (d) TDAC output. (e) TDAC's learned initialization map
$\phi_0(x,y)$ and parameter maps (f) $\lambda_{1}(x,y)$ and (g)
$\lambda_{2}(x,y)$.}
\label{fig:final_comparison}
\end{figure}

\begin{figure}[t] \centering
\def\x{0.135}
\includegraphics[width=\x\linewidth,height=\x\linewidth]{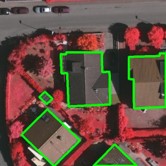}
\hfill
\includegraphics[width=\x\linewidth,height=\x\linewidth]{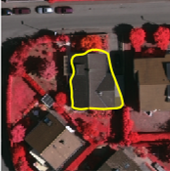}
\hfill
\includegraphics[width=\x\linewidth,height=\x\linewidth]{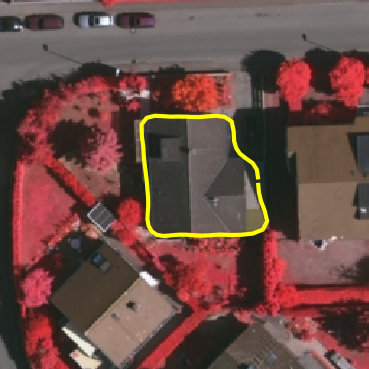}
\hfill
\includegraphics[width=\x\linewidth,height=\x\linewidth]{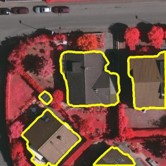}
\hfill
\includegraphics[width=\x\linewidth,height=\x\linewidth]{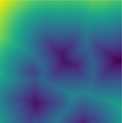} \hfill
\includegraphics[width=\x\linewidth,height=\x\linewidth]{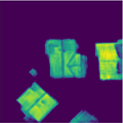} \hfill
\includegraphics[width=\x\linewidth,height=\x\linewidth]{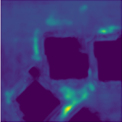}\\[4pt]

\includegraphics[width=\x\linewidth,height=\x\linewidth]{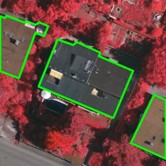}
\hfill
\includegraphics[width=\x\linewidth,height=\x\linewidth]{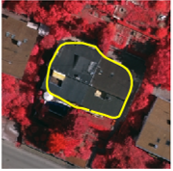}
\hfill
\includegraphics[width=\x\linewidth,height=\x\linewidth]{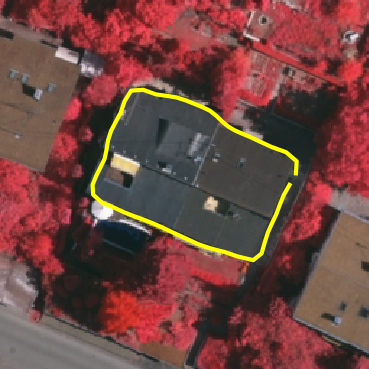}
\hfill
\includegraphics[width=\x\linewidth,height=\x\linewidth]{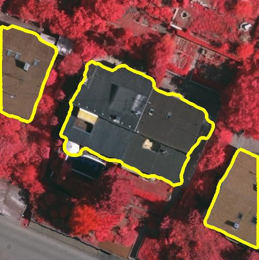}
\hfill
\includegraphics[width=\x\linewidth,height=\x\linewidth]{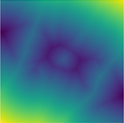} \hfill
\includegraphics[width=\x\linewidth,height=\x\linewidth]{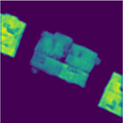} \hfill
\includegraphics[width=\x\linewidth,height=\x\linewidth]{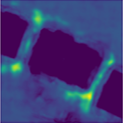}\\[4pt]

\includegraphics[width=\x\linewidth,height=\x\linewidth]{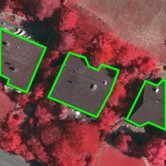}
\hfill
\includegraphics[width=\x\linewidth,height=\x\linewidth]{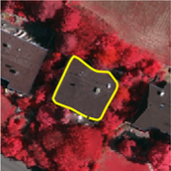}
\hfill
\includegraphics[width=\x\linewidth,height=\x\linewidth]{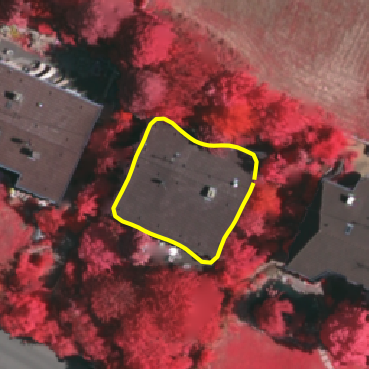}
\hfill
\includegraphics[width=\x\linewidth,height=\x\linewidth]{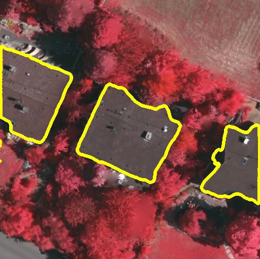}
\hfill
\includegraphics[width=\x\linewidth,height=\x\linewidth]{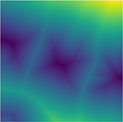} \hfill
\includegraphics[width=\x\linewidth,height=\x\linewidth]{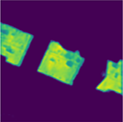} \hfill
\includegraphics[width=\x\linewidth,height=\x\linewidth]{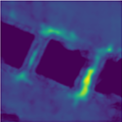}\\[4pt]

\begin{subfigure}{\textwidth} \centering
\subcaptionbox{Image}{\includegraphics[width=\x\linewidth,height=\x\linewidth]{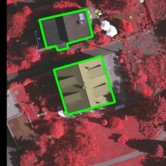}}
\hfill
\subcaptionbox{DSAC}{\includegraphics[width=\x\linewidth,height=\x\linewidth]{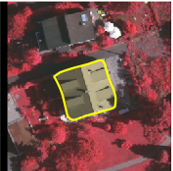}}
\hfill
\subcaptionbox{DarNet}{\includegraphics[width=\x\linewidth,height=\x\linewidth]{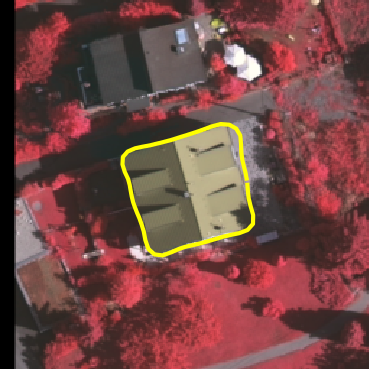}}
\hfill
\subcaptionbox{\bf TDAC}{\includegraphics[width=\x\linewidth,height=\x\linewidth]{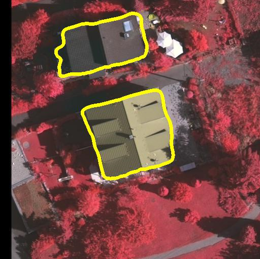}}
\hfill
\subcaptionbox{$\phi_0(x,y)$}{\includegraphics[width=\x\linewidth,height=\x\linewidth]{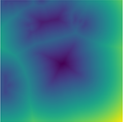}}
\hfill
\subcaptionbox{$\lambda_1(x,y)$}{\includegraphics[width=\x\linewidth,height=\x\linewidth]{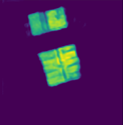}}
\hfill
\subcaptionbox{$\lambda_2(x,y)$}{\includegraphics[width=\x\linewidth,height=\x\linewidth]{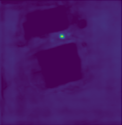}}
\end{subfigure}
\caption{Additional comparative visualization of the labeled image and
the outputs of DSAC, DarNet, and our TDAC, for the Vaihingen dataset.}
\label{fig:final_comparison1}
\end{figure}

\begin{table}[t] \centering
\setlength{\tabcolsep}{4pt}
\caption{Model Evaluations: Single-Instance Segmentation.}
\label{table:datasets-perf}
\resizebox{\columnwidth}{!}{%
\begin{tabular}{ll cccc cccc}
\toprule
\multicolumn{2}{c}{Model}  & \multicolumn{4}{c}{Vaihingen} & \multicolumn{4}{c}{Bing Huts} \\
   \cmidrule(lr){1-2} \cmidrule(lr){3-6} \cmidrule(lr){7-10}
Method    &   Backbone & Dice  &   mIoU &   WCov  &  BoundF   &   Dice  &   mIoU &   WCov  &  BoundF  \\
\midrule
FCN  & ResNet  & 84.20	& 75.60&	77.50&	38.30 & 79.90 &	68.40 &	76.14&	39.19 \\
FCN  & Mask R-CNN  & 86.00	&76.36 & 81.55 &	36.80 & 77.51& 65.03 &	76.02 &	65.25 \\
FCN    &  UNet  & 87.40 &	78.60 &	81.80 &	40.20 & 77.20 &	64.90&	75.70&	41.27  \\
FCN    &  Ours  & 90.02 &	81.10 &	82.01 &	44.53 & 82.24 &	74.09&	73.67&	42.04  \\
FCN &  DSAC  & -- &	81.00 &	81.40 &64.60	 & -- &69.80	&73.60	& 30.30	  \\
FCN &  DarNet  & -- &87.20	 &86.80 &	76.80 & -- &74.50	&77.50	& 37.70	  \\
DSAC &  DSAC  & -- &71.10	 &70.70	 &	36.40 & -- &38.70	&44.60	& 37.10	  \\
DSAC &  DarNet  & -- &60.30	 &61.10	 &	24.30 & -- &57.20	&63.00	& 15.90	  \\
DarNet &  DarNet  & 93.66 &88.20	 &88.10	 &	75.90 & 85.21 &75.20	&77.00	& 38.00	  \\
TDAC-const\,$\lambda$s  & Ours &  91.18 &	83.79 &  82.70 &	73.21 & 84.53 &	73.02 & 74.21 & 48.25	 \\
TDAC  & Ours&  \textbf{94.26}  &	\textbf{89.16}	& \textbf{90.54}   & \textbf{78.12} & \textbf{89.12} &	\textbf{80.39}   & \textbf{81.05}    &	\textbf{53.50}  \\
\bottomrule
\end{tabular}
}
\end{table}

\begin{table}[t] \centering
\setlength{\tabcolsep}{4pt}
\caption{Model Evaluations: Multiple-Instance Segmentation.}
\label{table:datasets-perf2}
\resizebox{\columnwidth}{!}{
\begin{tabular}{ll cccc cccc}
\toprule
\multicolumn{2}{c}{Model}  & \multicolumn{4}{c}{Vaihingen} & \multicolumn{4}{c}{Bing Huts} \\
\cmidrule(lr){1-2} \cmidrule(lr){3-6} \cmidrule(lr){7-10}
Method    &   Backbone & Dice  &   mIoU &   WCov  &  BoundF   &   Dice  &   mIoU &   WCov  &  BoundF  \\
\midrule
FCN     & UNet  &81.00&	69.10&	72.40&	34.20& 71.58&	58.70&	65.70&	40.60  \\ 
FCN     & ResNet  & 80.10 & 67.80   &	70.50&	32.50 & 74.20 &	61.80&	66.59&	39.48 \\ 
FCN     & Mask R-CNN  & 88.35 & 79.42   &	80.26 &	41.92 & 76.12 & 63.40 & 70.51 & 41.97 \\ 
FCN     & Ours & 89.30	& 81.00 &	82.70 &	49.80 & 75.23 &	60.31 & 72.41	 & 41.12	  \\
TDAC-const\,$\lambda$s  & Ours&  90.80	&83.30	&83.90	&47.20	  & 81.19 & 68.34 & 75.29 & 44.61 \\
{TDAC} & Ours&   \textbf{95.20} & \textbf{91.10} &	\textbf{91.71} &	\textbf{69.02} &  \textbf{83.24} &	\textbf{71.30}   & \textbf{78.45}    &	\textbf{48.49} \\
\bottomrule
\end{tabular}
}
\end{table}

\begin{figure}[t] \centering
\def\x{0.19}
\subcaptionbox{Image and\\green GT label}{\includegraphics[width=\x\linewidth,height=\x\linewidth]{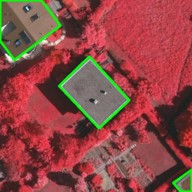}}\hfill
\subcaptionbox{TDAC with\\ constant
$\lambda_1$, $\lambda_2$}{\includegraphics[width=\x\linewidth,height=\x\linewidth]{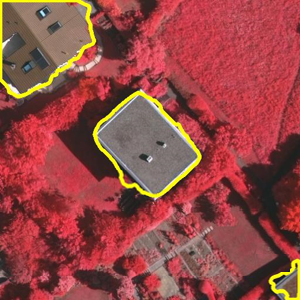}}\hfill
\subcaptionbox{TDAC}{\includegraphics[width=\x\linewidth,height=\x\linewidth]{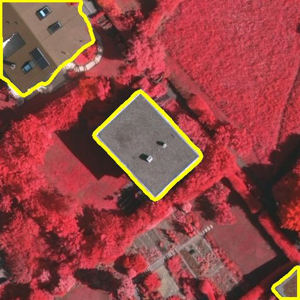}}\hfill
\subcaptionbox{$\lambda_1(x,y)$}{\includegraphics[width=\x\linewidth,height=\x\linewidth,trim={80
35 20 20},clip]{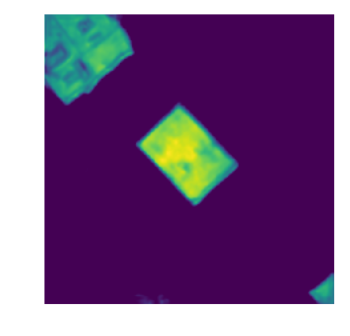}}\hfill
\subcaptionbox{$\lambda_2(x,y)$}{\includegraphics[width=\x\linewidth,height=\x\linewidth,trim={80
35 20 20},clip]{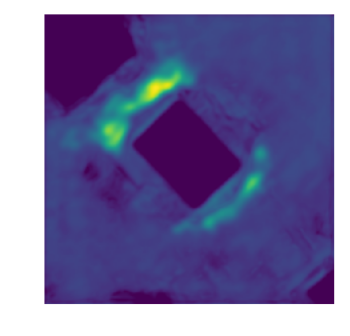}}
\caption{(a) Image labeled with (green) ground truth segmentation. (b)
Output of TDAC with constant $\lambda_1$ and $\lambda_2$. (c) TDAC
output and learned parameter maps (d) $\lambda_{1}(x,y)$ and (e)
$\lambda_{2}(x,y)$.}
\label{fig:const-lambda}
\end{figure}

\paragraph{Single-Instance Segmentation:}

Although most of the images in the Vaihingen dataset depict multiple
instances of buildings, the DarNet and DSAC models can deal only with
a single building instance at a time. For a fair comparison against
these models, we report single-instance segmentation results in the
exact same manner as~\cite{marcos2018learning}
and~\cite{cheng2019darnet}. As reported in
Table~\ref{table:datasets-perf}, our TDAC model outperforms both
DarNet and DSAC in all metrics on both the Vaihingen and Bing Huts
datasets. Fig.~\ref{fig:final_comparison} shows that with the
Vaihingen dataset, both the DarNet and DSAC models have difficulty
coping with the topological changes of the buildings and fail to
appropriately capture sharp edges, while TDAC overcomes these
challenges in most cases. For the Bing Huts dataset, both the DarNet
and DSAC models are able to localize the buildings, but they
inaccurately delineate the buildings in many cases. This may be due to
their inability to distinguish the building from the surrounding
terrain because of the low contrast and small size of the image.
Comparing the segmentation output of DSAC
(Fig.~\ref{fig:final_comparison}b), DarNet
(Fig.~\ref{fig:final_comparison}c), and TDAC
(Fig.~\ref{fig:final_comparison}d), our model performs well on the low
contrast dataset, delineating buildings more accurately than the
earlier models. Additional comparative visualizations are presented in
Fig.~\ref{fig:final_comparison1}.

\paragraph{Multiple-Instance Segmentation:}

We next compare the performance of TDAC against popular models such as
Mask R-CNN for multiple-instance segmentation of all buildings in the
Vaihingen and Bing Huts datasets. As reported in
Table~\ref{table:datasets-perf2}, our extensive benchmarks confirm
that the TDAC model outperforms Mask R-CNN and the other
methods by a wide margin. Although Mask R-CNN seems to be able
to localize the building instances well, the fine-grained details of
boundaries are lost, as is attested by the BoundF metric. The performance
of other CNN-based approaches follow the same trend in our benchmarks.

\paragraph{Parameter Maps:}

To validate the contribution of the parameter maps $\lambda_1(x,y)$ and $\lambda_2(x,y)$ in the
level-set ACM, we also trained our TDAC model on both the Vaihingen and Bing
Huts datasets by allowing just two trainable scalar parameters,
$\lambda_1$ and $\lambda_2$, constant over the entire image. As
reported in Table~\ref{table:datasets-perf}, for both the Vaihingen
and Bing Huts datasets, this ``constant-$\lambda$'' formulation (i.e.,
the Chan-Vese model~\cite{chan2001active,lankton2008localizing}) 
still outperforms the baseline CNN in most evaluation metrics for both
single-instance and multiple-instance buildings, thus establishing the
effectiveness of the end-to-end training of the TDAC. Nevertheless,
our TDAC with its full $\lambda_1(x,y)$ and $\lambda_2(x,y)$ maps
outperforms this constant-$\lambda$ version by a wide margin in all
experiments and metrics. A key metric of interest in this comparison
is the BoundF score, which elucidates that our formulation
captures the details of the boundaries more effectively by locally
adjusting the inward and outward forces on the contour.
Fig.~\ref{fig:const-lambda} shows that our TDAC has well delineated
the boundaries of the building instances, compared to the TDAC hobbled
by the constant-$\lambda$ formulation.

\paragraph{Convolutional Filter Size:}

\begin{figure}[t] \centering
\subcaptionbox{}{\includegraphics[width=0.42\linewidth,height=0.32\linewidth,trim={0 0 0 40},clip]{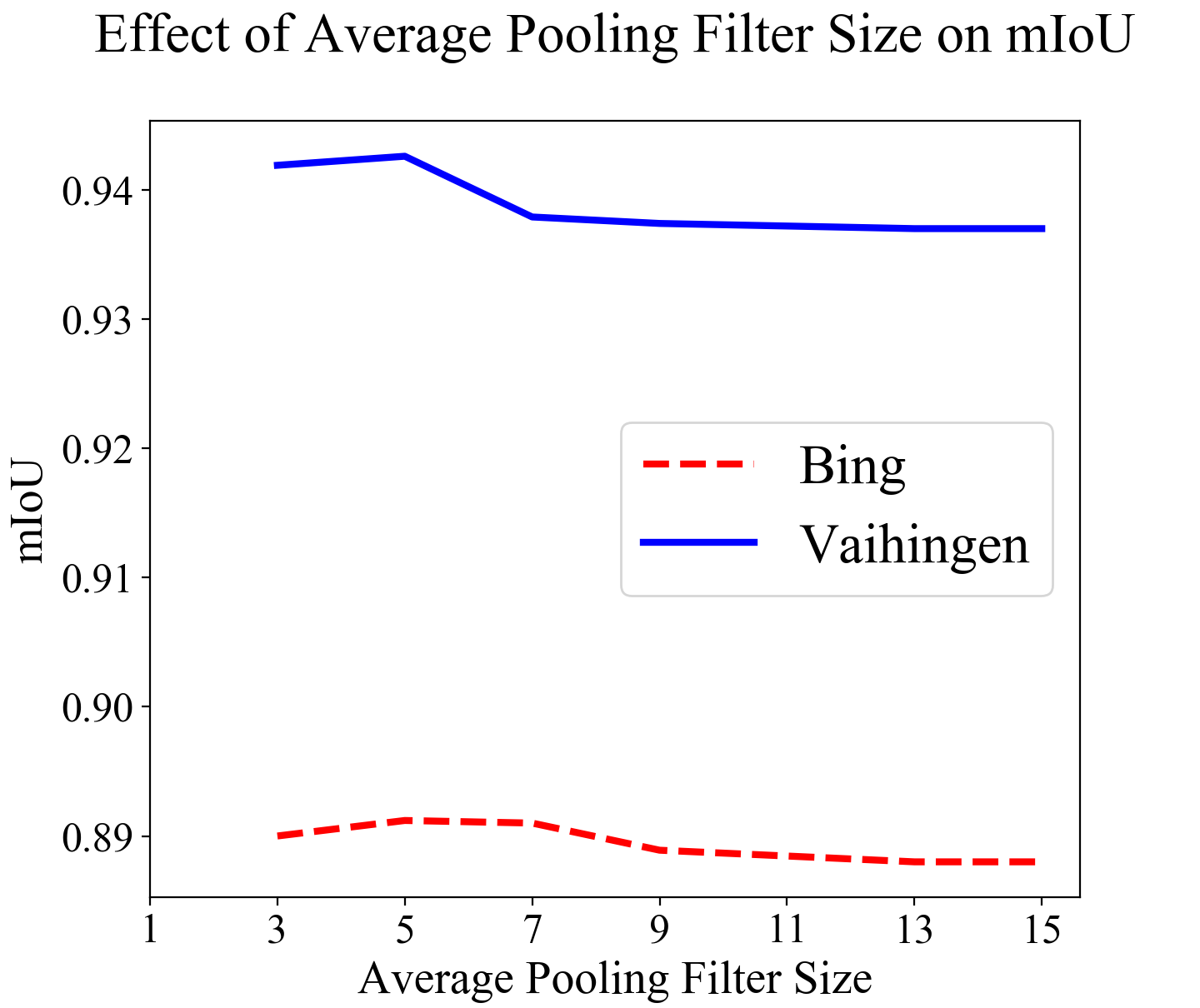}} 
\qquad
\subcaptionbox{}{\includegraphics[width=0.42\linewidth,height=0.32\linewidth,trim={0 0 0 40},clip]{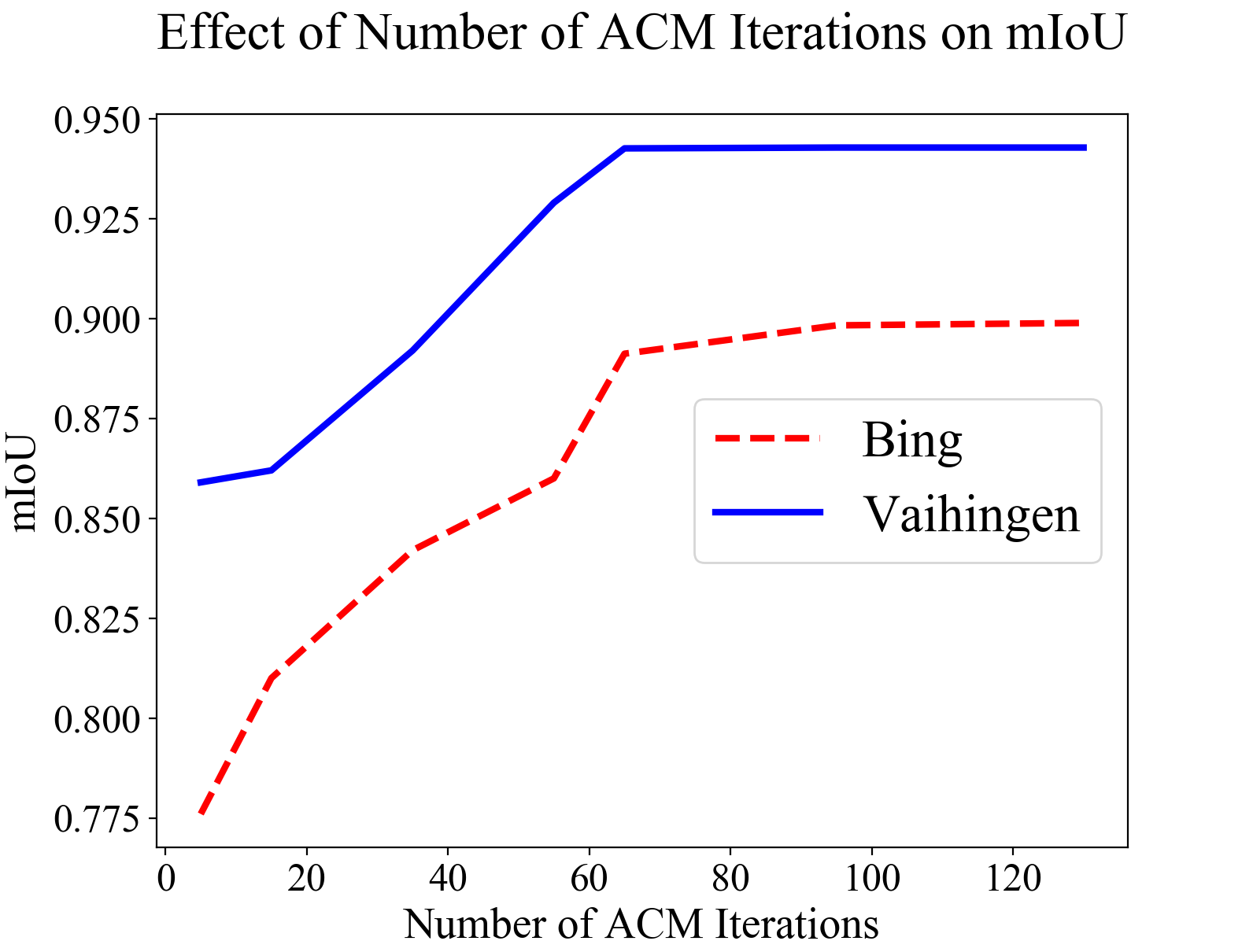}}
\caption{The effects on mIoU of (a) varying the convolutional filter size and 
(b) varying the number $L$ of ACM iterations.}
  \label{fig:exp_plots}
\end{figure}

The filter size of the convolutional operation is an important
hyper-parameter for the accurate extraction of localized image
statistics. As illustrated in Fig.~\ref{fig:exp_plots}a, we have
investigated the effect of the convolutional filter size on the
overall mIoU for both the Vaihingen and Bing datasets. Our experiments
indicate that filter sizes that are too small are sub-optimal while
excessively large sizes defeat the benefits of the localized
formulation. Hence, we recommend a filter size of $f=5$ for the TDAC.

\paragraph{Number of Iterations:}

The direct learning of an initialization map $\phi_0(x,y)$ as well as its efficient TensorFlow implementation have enabled the TDAC to require substantially
fewer iterations to converge with a better chance of
avoiding undesirable local minima. As shown in
Fig.~\ref{fig:exp_plots}b, we have investigated the effect
of the number of iterations on the overall mIoU for both Vaihingen and
Bing datasets and our results reveal that TDAC exhibits a robust
performance after a certain threshold. Therefore, we have chosen a
fixed number of iterations (i.e., ACM layers) for optimal performance, $L=60$, yielding a runtime of less than 1\,sec in TensorFlow.

\section{Conclusions and Future Work}
\label{sec:conc}

We have introduced a novel image segmentation framework, called Trainable Deep Active Contour Models (TDACs),
which is a full, end-to-end merger of ACMs and CNNs. To this end, we proposed a new,
locally-parameterized, Eulerian ACM energy model that includes pixel-wise
learnable parameter maps that can adjust the contour to precisely delineate the boundaries of objects of interest in the image. Our model is
fully automatic, as its backbone CNN learns the ACM initialization map as
well as the parameter maps that guide the contour to avoid suboptimal
solutions. This eliminates any reliance on manual
initialization of ACMs. Moreover, by contrast to previous approaches that have
attempted to combine CNNs with ACMs that are limited to segmenting a
single building at a time, our TDAC can segment any number of
buildings simultaneously.

We have tackled the problem of building instance segmentation on two
challenging datasets, Vaihingen and Bing Huts, and our
model significantly outperforms the current state-of-the-art methods on these test cases.

Given the level of success that TDAC has achieved in the building delineation application
and the fact that it features an Eulerian ACM formulation, it is
readily applicable to other segmentation tasks in various domains,
wherever purely CNN filter-based approaches can benefit from the
versatility and precision of ACMs to accurately delineate object boundaries in
images.

\appendix

\section{Derivation of the ACM Evolution PDE}

Following~\cite{lankton2008localizing}, we derive the Euler-Lagrange
PDE governing the evolution of the ACM. Let $X_{1}=(u,v)$ and
$X_{2}=(x,y)$ represent two independent spatial variables that each
represent a point in $\Omega$. Using the characteristic function
(\ref{eq:charf}), which selects regions within a square window of size
$s$, the energy functional of $C$ may be written in terms of a generic
internal energy density $F$ as
\begin{equation}
\label{eq:f_pc_region_final}
E(\phi)= \int_{\Omega_{X_{1}}}
\delta(\phi(X_{1}))\int_{\Omega_{X_{2}}} W_s
F(\phi,X_{1},X_{2})\,dX_{2}\,dX_{1}.
\end{equation}
To compute the first variation of the energy functional, we add to
$\phi$ a perturbation function $\epsilon\psi$, where $\epsilon$ is a
small number; hence,
\begin{equation}
\label{eq:f_pc_region_purt}
E(\phi+\epsilon\psi)= \int_{\Omega_{X_{1}}}
\delta(\phi(X_{1})+\epsilon\psi)\int_{\Omega_{X_{2}}} W_s
F(\phi+\epsilon\psi,X_{1},X_{2})\,dX_{2}\,dX_{1}.
\end{equation}
Taking the partial derivative of (\ref{eq:f_pc_region_purt}) with
respect to $\epsilon$ and evaluating at $\epsilon=0$ yields, according
to the product rule,
\begin{equation}
\label{eq:f_pc_derv}
\begin{split}
\left.\dfrac{\partial E}{\partial\epsilon}\right|_{\epsilon=0} =
\int_{\Omega_{X_{1}}} \delta(\phi(X_{1}))\int_{\Omega_{X_{2}}}\psi W_s
\nabla_{\phi}F(\phi,X_{1},X_{2})\,dX_{2}\,dX_{1} +
\\ \psi\int_{\Omega_{X_{1}}}\gamma\phi(X_{1})\int_{\Omega_{X_{2}}} W_s
F(\phi,X_{1},X_{2})\,dX_{2}\,dX_{1},
\end{split}
\end{equation}
where $\gamma\phi$ is the derivative of $\delta(\phi)$. Since
$\gamma\phi$ is zero on the zero level set, it does not affect the
movement of the curve. Thus the second term in (\ref{eq:f_pc_derv})
and can be ignored. Exchanging the order of integration, we obtain
\begin{equation}
\label{eq:f_pc_derv_v1}
\left.\dfrac{\partial E}{\partial\epsilon}\right|_{\epsilon=0} =
\int_{\Omega_{X_{2}}}\int_{\Omega_{X_{1}}} \psi\delta(\phi(X_{1})) W_s
\nabla_{\phi}F(\phi,X_{1},X_{2})\,dX_{1}\,dX_{2}.
\end{equation}
Invoking the Cauchy–Schwartz inequality yields
\begin{equation}
\label{eq:evol_region_v2}
\frac{\partial\phi}{\partial t} =
\int_{\Omega_{X_{2}}}\delta(\phi(X_{1})) W_s
\nabla_{\phi}F(\phi,X_{1},X_{2})\,dX_{2}.
\end{equation}
Adding the contribution of the curvature term and expressing the
spatial variables by their coordinates, we obtain the desired curve
evolution PDE (\ref{eq:ACWEalgorithm}) with
(\ref{eq:final_evol}) under the assumption of a uniform internal
energy model with $m_1(x,y)$ and $m_2(x,y)$ as the mean image
intensities inside and outside $C$ and within $W_s$.

\bibliographystyle{splncs04}
\bibliography{eccv20}
\end{document}